\pdfoutput=1
\PassOptionsToPackage{table,xcdraw}{xcolor}
\documentclass[11pt]{article}

\usepackage[]{emnlp2021}

\usepackage{times}
\usepackage{latexsym}

\usepackage[T1]{fontenc}

\usepackage[utf8]{inputenc}

\usepackage{microtype}
\usepackage{accents}
\usepackage[table,xcdraw]{xcolor}
\usepackage{algorithmic}
\usepackage{longtable}
\usepackage[ruled,vlined]{algorithm2e}
\usepackage{float}
\usepackage{color}
\usepackage[singlelinecheck=false,justification=justified]{caption}
\usepackage{subcaption}
\usepackage[multiple]{footmisc}

\usepackage[colorinlistoftodos,prependcaption]{todonotes}

\newcommand{\tb}[1]{\textcolor{brown}{ #1}}

\usepackage{graphicx}

\title{Question Answering over Electronic Devices: A New Benchmark Dataset and a Multi-Task Learning based QA Framework}

\author{

Abhilash Nandy$^{\spadesuit}$ ~~~~
Soumya Sharma$^{\spadesuit}$ ~~~~
Shubham Maddhashiya$^{\spadesuit}$ ~~~~
Kapil Sachdeva$^\clubsuit$\\
\textbf{Pawan Goyal}$^{\spadesuit}$ ~
\textbf{Niloy Ganguly}$^{\spadesuit\diamondsuit}$\\ 
\\
$^\spadesuit$Indian Institute of Technology, Kharagpur ~
$^{\clubsuit}$Samsung Research Institute, Delhi  \\
$^{\diamondsuit}$ L3S Research Center, Leibniz Universität Hannover \\
}

\begin{document}
\maketitle
\begin{abstract}
Answering questions asked from instructional corpora such as E-manuals, recipe books, etc., has been far less studied than open-domain factoid context-based question answering. 
This can be primarily attributed to the absence of standard benchmark datasets. In this paper we meticulously create a large amount of data connected with E-manuals and develop suitable algorithm to exploit it. We collect \textbf{E-Manual Corpus}, a huge corpus of 307,957 E-manuals and pretrain RoBERTa on this large corpus. We create various benchmark QA datasets which include question answer pairs curated by experts based upon two E-manuals,  real user questions from Community Question Answering Forum pertaining to E-manuals etc. 
We introduce {\bf EMQAP (E-Manual Question Answering Pipeline)} 
that answers questions pertaining to  electronics devices.  
Built upon the pretrained RoBERTa, it harbors a  
 supervised multi-task learning framework which efficiently performs the dual tasks of identifying the section in the E-manual where the answer can be found and the exact answer span within that section. 
For E-Manual annotated question-answer pairs, we show an improvement of about 40\% in ROUGE-L F1 scores over the most competitive baseline. We perform a detailed ablation study and establish the versatility of EMQAP across different circumstances. The code and datasets are shared at~\url{https://github.com/abhi1nandy2/EMNLP-2021-Findings}, and the corresponding project website is~\url{https://sites.google.com/view/emanualqa/home}.
\end{abstract}

\section{Introduction}
An E-Manual, or Electronic Manual, is a document that provides technical support to the consumers of a product by giving instructions and procedures to operate the device along with know-how of its specifications. It is often difficult to find the relevant instructions from an E-manual; hence, an automated question answering support to use the information present in the E-manual effectively would be of great help. 
 
E-Manuals typically provide lengthy instructions structured in a sequential fashion explaining various uses of a device. This often poses a challenge in building a question answering system because the answer to a question may come from multiple disjointed portions within a section of the E-Manual.  
Due to the instructional nature of E-Manuals, we also find that often adjacent instructions are not related to each other but may be related to a parental instruction leading to long-range dependencies in context. This, therefore, deems a {\bf domain-specific natural language understanding} which may, in turn, suffer from lack of domain-specific labeled data ~\cite{finbert} and presence of formal syntax in the corpus \cite{scibert,legalbert}. These challenges have led recent works to pre-train the state-of-the-art transformer models on unlabelled domain-specific corpora \cite{biobert,finbert,scibert,legalbert}. Inspired by such works,  we painstakingly collect {\bf E-Manual Corpus: a huge corpus of 307,957 E-manuals}\footnote{\url{www.manualsonline.com}} 
and pre-train the transformer-based language model, RoBERTa\_{BASE}\footnote{Note that, in this paper, unless otherwise specified, `RoBERTa' would just mean `RoBERTa\_{BASE}'} on the corpus (Section \ref{pretrain-strategies}).  
A {\bf question answering system} needs to 
select the relevant section of the E-Manual, which contains the answer to the given question ({\bf section retrieval (SR)}) and subsequently, 
extract the answer from that relevant section ({\bf answer retrieval (AR)}). There are currently four main types of approaches in state-of-the-art literature that utilize the SR and AR systems (1). ~\newcite{twostage} uses a two-stage training pipeline where the SR model consists of an unsupervised Information Retrieval (IR) method like TF-IDF or BM25, followed by an extractive AR model; (2) an end-to-end learning setup of SR cascaded by AR  ~\cite{realm,end2end}; (3) single-span \cite{squad} or multi-span~\cite{long_multiple_spans, simple_multi_span} answers given questions and corresponding candidate contexts as inputs and (4) a Multi-task Learning (MTL) Framework, where SR and AR are the two underlying tasks ~\cite{retrieveread};  \newcite{retrieveread} performs MTL using separate SR and AR pipelines sharing feature extraction layers. The simultaneous training of SR and AR using MTL helps the model build a combined and hierarchical understanding of Question Answering at a global (section) and a local (sentence/token) level. However, these methods apply a span-based selection approach for extracting answers, whereas the answers to questions on E-Manuals are usually non-contiguous; hence while we principally use  this {\bf multi-task learning (MTL)} framework, we make some customization to accommodate the peculiarity of the data. 

Summing up, the paper makes the following contributions: 
\textbf{(1)} Since no data is available for the E-Manual domain, we create a huge corpus for pretraining containing 307,957 E-Manuals known as the E-Manual Corpus.
\textbf{(2)} Since no QA dataset is available for this domain, we apply multi-pronged strategy to create  a large enough corpus of  Question Answering (QA) datasets: two datasets {\bf manually annotated by experts} containing {\bf 904 and 950 questions} respectively, and another collected from {\bf Amazon Question Answering Forum containing 1,028 questions} and a set of {\bf 10 question-answer pairs for 40 different devices each} (Section \ref{corpora-and-datasets}).
\textbf{(3)} {EMQAP (E-Manual Question Answering Pipeline)} develops on two basic pillars -  a domain-specific {\bf pre-trained RoBERTa architecture} and a {\bf multi-task learning framework}. 

In the next section we discuss in detail the different types of data rigorously created. 
The system design is discussed in detail in Section \ref{Methodology},  followed by the experimental results
in Section \ref{experiments-results}.
The experimental results emphatically establish that the performance of EMQAP is way superior to its nearest baseline. 


    

\if{0} In this paper, we make the following contributions - (1) Exploring various methods and strategies of pre-training on a domain-specific corpus of E-Manuals. (2) coming up with a novel transformer-based multi-task learning architecture. (3) Analyzing the difficulty of answering annotated questions and real-life questions asked by consumers. \fi 

\section{Corpus and Datasets}
\label{corpora-and-datasets}
In this section, we  elaborate the corpus of E-Manuals and the benchmark datasets  we create. These datasets  are used for pre-training and to test the performance  of the QA algorithms. 
\subsection{Creating the  corpus of E-Manuals used for pre-training}\label{emanuals-corpus}
To perform pre-training, we create a large text corpus of E-Manuals by collecting and pre-processing \tb{(details in {\em suppl.})} text from $307,957$ pdf files downloaded from source\footnote{\url{www.manualsonline.com}}. All these pdf files serve as manuals for several categories of products and services, such as baby care, kitchen appliances, electronic goods, personal care, lawn, garden, etc. The variety prevents over-fitting to the E-Manuals of a specific product type. The details of the dataset have been summarized in Table~\ref{tab:pre-train-corpus}.
On plotting the  word cloud \tb{(figure in {\em suppl.})} for the most frequently occurring terms, it is found that words that make sentences instructional and assertive  e,g,. "avoid", "help", "handle", "leave", "print" are prominent .

\begin{table}[H]
\centering
\begin{tabular}{l|c}
\multicolumn{1}{c|}{\textbf{Property}} & \textbf{Value}  \\ \hline
No. of E-Manuals                    & 307,957         \\
No. of paragraphs                   & 11,653,755      \\
No. of sentences per paragraph      & 4.4             \\
No. of words per sentence           & 20.2            \\
Total number of words                  & $\sim$1 Billion \\
Size of corpus (in GB)                 & $\sim$11 GB    
\end{tabular}
\caption{Details of the E-Manual pre-training corpus used in terms of property-value pairs}
\label{tab:pre-train-corpus}
\end{table}




\subsection*{Question Answering Dataset}

We create datasets of different types  which can act as   benchmarks to test the performance of a E-Manual Question Answering  algorithm   under  varied  circumstances. We consider two most popular categories of consumer items, mobile and smart TV. For each of these categories, we take a representative E-manual and employ experts to curate questions covering all sections of these manuals.  We also check what are the questions raised by smart TV users on online forums. Finally, we expand our domain to 40 devices of different categories and collect a small representative QA for them to check the versatility of the algorithm. 
For all our datasets, we decided to choose a single brand to have some sort of consistency across E-manuals, incidentally we chose `Samsung' due to convenience \tb{(reasons detailed in {\em suppl.})}. 
However, other popular brands could also be chosen, we believe that would not make much of a difference.  
Note, except for TechQA Dataset~\cite{techqa} which is built from questions regarding general software based technical support and hardly contains any question pertaining to E-manual, to the best of our knowledge, no such similar dataset is  available. 


\subsection{Question Answering Dataset from E-Manual}
\label{qna-datasets}
We have selected E-Manual of a Samsung S10 phone~\cite{s10-manual}and a Samsung Smart TV/remote~\cite{Tv-remote-manual} and created corresponding question-answer datasets with the help of expert annotators.   
Each section is carefully read by an annotator~\footnote{\url{http://www.tika-data.com/}} and she has accordingly posed questions and  marked certain sentences from the section as the answer.   An E-Manual’s sections were split among $3$ annotators to reduce cognitive load. The annotators were non-native but fluent English speakers. 
Annotators also curated   {\bf paraphrased questions}  where an already existing question is expressed differently, e.g., "How do I turn off sound notifications?" is paraphrased as "How can I mute all notification sound?". A crowdsource based quality assessment of the  annotations is conducted  \tb{({\em detail in suppl})} and is found to be satisfactory.
The stats of our datasets along with the TechQA Dataset~\cite{techqa} are presented  in Table~\ref{tab:datasets}. 

\begin{table*}[t]
\centering
\tiny
\begin{tabular}{l|lcccccccl}
\multicolumn{1}{c|}{\textbf{Dataset}} &
  \multicolumn{1}{c}{\textbf{Domain}} &
  \textbf{\begin{tabular}[c]{@{}c@{}}No. of\\QA pairs\end{tabular}} &
  \textbf{\begin{tabular}[c]{@{}c@{}c@{}}\%age of\\factual\\questions\end{tabular}} &
  \textbf{\begin{tabular}[c]{@{}c@{}c@{}}\%age of\\procedural\\questions\end{tabular}} &
  \textbf{\begin{tabular}[c]{@{}c@{}c@{}}\%age of\\questions asking\\feature location\end{tabular}} &  
  \textbf{\begin{tabular}[c]{@{}c@{}c@{}}\%age of\\paraphrased\\questions\end{tabular}} &  
  \textbf{\begin{tabular}[c]{@{}c@{}c@{}}Avg\\Question\\Length\end{tabular}} &
  \textbf{\begin{tabular}[c]{@{}c@{}c@{}}Avg.\\Answer\\Length\end{tabular}} &
  \textbf{\begin{tabular}[c]{@{}c@{}}Answer\\Type\end{tabular}} \\ 
  \hline
\begin{tabular}[c]{@{}l@{}}TechQA\\ \cite{techqa}\end{tabular} &
  \begin{tabular}[c]{@{}l@{}}Technical\\ Support\end{tabular} &
  1,400 &
  22.75 &
  32.64 &
  0.88 &
  0 &
  52.5 &
  45 &
  \begin{tabular}[c]{@{}l@{}}Single Span,\\ long answer\end{tabular} \\ \hline  
\begin{tabular}[c]{@{}l@{}}S10 QA\end{tabular} &
  E-Manual &
  904 &
  7.08 &
  48.34 &
  7.3 &
  33.52 &  
  9.4 &
  48.4 &   
  \begin{tabular}[c]{@{}l@{}}Multi Span,\\ long answer\end{tabular} \\ \hline
\begin{tabular}[c]{@{}l@{}}Smart TV/Remote QA\end{tabular} &
  \begin{tabular}[c]{@{}l@{}}E-Manual\end{tabular} &
  950 &
  14.26 &
  51.74 &
  3.03 &
  30.35 &
  11 &
  61.5 &   
  \begin{tabular}[c]{@{}l@{}}Multi Span,\\ long answer\end{tabular} \\ \hline
\begin{tabular}[c]{@{}l@{}}Smart TV/Remote\\ Amazon Consumer Questions\end{tabular} &
  \begin{tabular}[c]{@{}l@{}}User\\ Forum\end{tabular} &
  1,028 &
  12.35 &
  37.06 &
  0.97 &
  0 &
  12.84 &
  20.41 &   
  \begin{tabular}[c]{@{}l@{}}Multi Span,\\ long answer\end{tabular} 
\end{tabular}
\caption{Description of our datasets and the TechQA Dataset. The \% showing various categories (including the paraphrase) does not sum upto to 100 as some questions cannot be classified into one of the three categories. The categories of the paraphrase is not shown as they roughly follow the similar distribution of the unique questions.}
\vspace{-6mm}
\label{tab:datasets}
\end{table*}

\begin{figure}[H]
    \centering
    \includegraphics[scale=0.9]{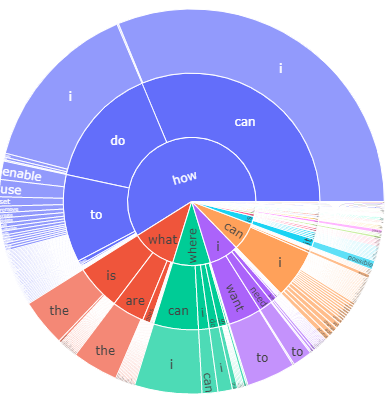}
    \caption{Distribution of questions covered in S10 QA Dataset w.r.t their first three tokens.}
    \label{fig:q_distr}
\end{figure}

Most of the questions belong to one of these three categories - (a). about facts regarding device operations, which we refer to as ``Factual". (`what', `which', `why', `when' type questions) (b). on \textit{how} to carry out a specific operation referred as ``Procedural" (`how', `can' type questions) (c) asking the location of a particular feature (`where' type questions).
We show 
the distribution of questions w.r.t the first three tokens for Samsung S10 in Fig.~\ref{fig:q_distr}. It shows that more than $50\%$ of the questions are `how' type questions (`how can', `how to' etc.), while `what', `where' and `can' type questions also have a significant percentage. There are also a few questions, which start with `I want to', `I need to', which start with the end user's desired functionality followed by a question (``I want to switch on Bluetooth. What should I do?'').

\subsection{Questions from the real consumers}
\label{reallife}

The QA dataset of the Samsung Smart TV manual is used to sanitize a community-based question answering dataset described next. 
Questions are extracted from question answering forum (where well-formed answers are available) of the different Samsung Smart TV models sold on amazon. 
Annotators are asked to certify 
whether a question is answerable by solely using the E-Manual of the product. The dataset has a total of $3,000$ such questions, out of which $1,028$ are certified as answerable.  
Also, for each question, they were asked to select the most similar question from the manually annotated QA dataset created for Samsung Smart TV/Remote.  This would provide paraphrases for the relevant Consumer Questions, and the Consumer Question-Annotated Question pairs so formed are referred to as the CQ-AQ Dataset. The CQ-AQ Dataset covers $312$ of the  annotated answers in the Smart TV/Remote QA Dataset, hence have the answer from the e-manual as the \textbf{ANNotated-Ground Truth (ANN-GT)}. The other Ground Truth for a CQ-AQ pair is the answer from the Amazon Community Question Answering (CQA) Forum corresponding to the CQ, which is the CQA-Ground Truth (CQA-GT). 
We thus create a dataset consisting of $1028$ tuples, where each tuple consists of  [CQ-AQ, ANN-GT, CQA-GT].  


\subsection{Questions spanning across several devices}
\label{dataset_other_emanuals}

In this step, we curate $10$ generic Question-Answer pairs for $40$ devices on Amazon~\footnote{$13$ Samsung Galaxy Mobile Phones, $9$ other Samsung Mobile Phones, $15$ Samsung Tablets and $3$ Samsung Smart Watches}.  We sample $10$ questions from the S10 QA Dataset that would apply to a broad suite of devices. These $10$ questions are sampled so that their corresponding annotated answers are from different sections of the E-Manual, and $1$ is factual, $8$ are procedural, and $1$ is asking the location of a feature. These $10$  questions are \tb{{\em  listed in suppl.}} 
We consider 40 devices of different types. 
For each device, for each of the $10$ sample questions, the most relevant question is selected from the Amazon QA for that device using the CQ-AQ Paraphrase Detector \tb{\textit{discussed in suppl.}}  The answer corresponding to each question from Amazon is taken as the ground truth answer. Thus, we have 10 question pairs and a corresponding set of $10$ answers as the dataset for each of the $40$ devices. 

\section{Methodology}
\label{Methodology}

\begin{figure*}[t]
	\centering
	\subfloat[EMQAP: Pre-training and Multi-Task Learning]{%
		\centering 
\includegraphics[scale=0.32]{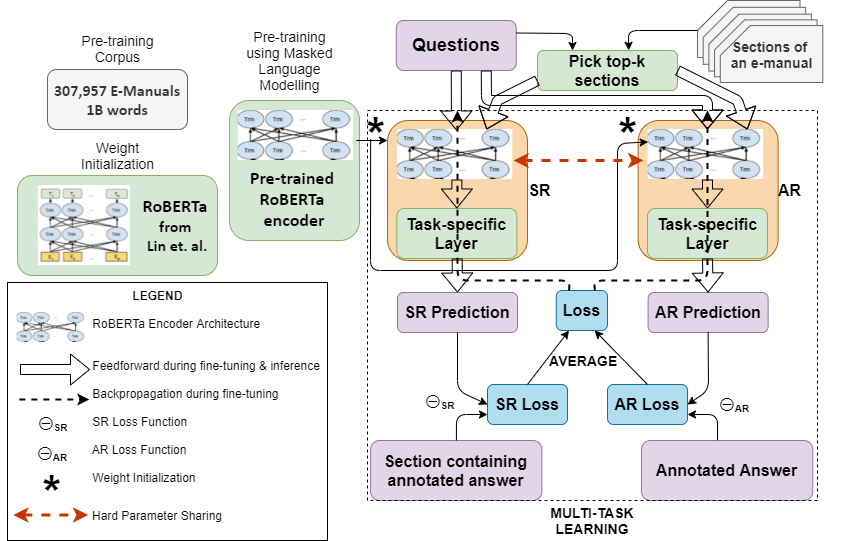} \label{1attrev}}
	\subfloat[Inference of EMQAP]{%
		\centering 
		\includegraphics[scale=0.32]{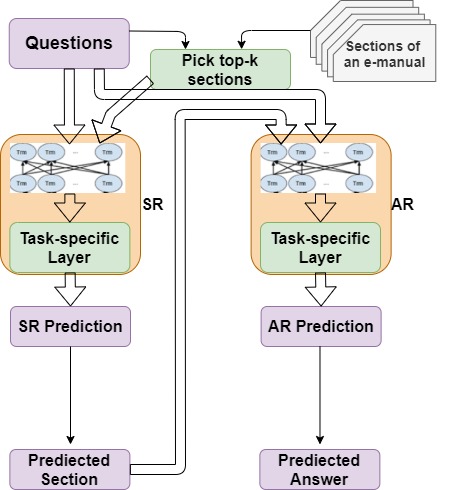} \label{1attadd}}\\
	\caption{EMQAP: RoBERTa architecture is used for pre-training with E-manuals, and its weights are used to initialize the SR and AR models of the MTL framework. A question along with the top $K$ relevant sections 
	form inputs to the SR and AR modules of the MTL Framework during training, and an average of the AR and SR losses is backpropagated through the whole framework. {\bf During inference}, once top-$k$ sections are retrieved from the unsupervised IR, the SR module outputs the most relevant section for the question; the question along with this predicted section are sent as input to the AR module, which finally predicts the answer to the question.}
	\label{fig:pipeline}
\end{figure*}

In this section, we describe each step from the pipeline of {\bf EMQAP}.  
The pipeline consists of two major
steps (a). {\bf pre-training} the E-manual and (b). {\bf multi-task learning} framework to select the answer. 
However, before employing multi-task learning, the first step is to reduce the pipeline's search space and provide it with only a few candidate sections for a question. We use an {\bf unsupervised IR} method that accepts a question and all sections of the E-Manual as input and provides similarity scores for each question-section as output \tb{({\em details in suppl.})}
The flow of the entire EMQAP is depicted in Fig. \ref{fig:pipeline}. 
 The steps are also \tb{presented} as  Algorithm  \tb{in {\em suppl}.}. 

 \subsection{Pre-training on the E-Manuals corpus}\label{pretrain-strategies} 
 A huge corpus of E-Manuals is used to pre-train the RoBERTa transformer using masked language modeling by masking $15\%$ of the tokens in each input string to enhance the domain-specific knowledge of our language model. Note, the base "RoBERTa" transformer architecture is already initialized by weights obtained by pre-training it on Wikipedia, and BooksCorpus  ~\cite{roberta}. 

We apply the following two pre-training strategies to efficiently capture both the generic and domain-specific knowledge required to answer a question. 
(a).  Using a learning rate that linearly decreases by a constant factor (LRD) from one layer to the next, with the outermost language modeling head layer having the maximum learning rate, as in \newcite{empiricalmodelswitch}. This 
enforces a constraint that outer layers adapt more to the E-Manual domain, while the inner layers' weights do not change much, thus restricting them to retain the knowledge of the generic domain primarily. (b). Using elastic weight consolidation (EWC) \cite{ewc,empiricalmodelswitch} to mitigate catastrophic forgetting while switching from the generic domain on which original "RoBERTa" was pre-trained to the domain of E-Manuals.  
A batch size of $64$ is used. Since our corpus size (11GB) is quite small compared to the datasets used for pre-training in \newcite{roberta}, we use a smaller batch size than used in \newcite{roberta}. However, the number of tokens per sentence is $20.2$, which ensures that a batch has a large number of tokens even with a smaller batch size. 
We pre-train for $1$ epoch since the training loss reaches a plateau, and does not reduce further at the end of the epoch. More details and justification for choosing the above mentioned techniques are \tb{{\em detailed in suppl.}} 

We wanted to have a subjective analysis as to how pre-training  helped the model learn better domain-specific context. 
We compared the model with off-the-shelf RoBERTa Model. Top $100$ most frequent words (excluding stopwords and numbers) present in the first $100,000$ lines of the EManuals Corpus are taken. 
For each word, top $5$ neighbours (based on cosine distance) are calculated for each model. The word and its neighbours are much more contextually related (through manual analysis) in case of  RoBERTa pretrained on E-Manuals, showing that, pre-training on E-Manuals enhances the context and meaning of domain-specific words.
$10$ such samples are shown in Table~\ref{tab:word_embeds}.


\begin{table}[H]
\centering
\resizebox{0.5\textwidth}{!}{%
\begin{tabular}{l|l|l} \hline
\textbf{Word} &
  \textbf{\begin{tabular}[c]{@{}l@{}}Top 5 nearest neighbours \\ for RoBERTa\end{tabular}} &
  \textbf{\begin{tabular}[c]{@{}l@{}}Top 5 nearest neighbours for \\ RoBERTa pre-trained \\ on E-Manuals\end{tabular}} \\ \hline
key &
  \textbf{button}, ip, must, field, note &
  \begin{tabular}[c]{@{}l@{}}\textbf{press}, note, \textbf{click}, \textbf{button}, \\ parameter\end{tabular} \\
address &
  \begin{tabular}[c]{@{}l@{}}support, phone, message,\\ button, change\end{tabular} &
  \begin{tabular}[c]{@{}l@{}}name, \textbf{server}, \textbf{message}, \\ \textbf{network}, local\end{tabular} \\
port &
  \begin{tabular}[c]{@{}l@{}}operation, enabled, must, \\ unit, enable\end{tabular} &
  \textbf{ports}, \textbf{ip}, \textbf{server}, \textbf{device}, unit \\
support &
  \begin{tabular}[c]{@{}l@{}}control, description, address, \\ ports, settings\end{tabular} &
  \begin{tabular}[c]{@{}l@{}}\textbf{information}, \textbf{service}, \textbf{call},\\  \textbf{3com}, \textbf{web}\end{tabular} \\
switch &
  \begin{tabular}[c]{@{}l@{}}operation, \textbf{change}, enabled,\\  unit, \textbf{button}\end{tabular} &
  \begin{tabular}[c]{@{}l@{}}\textbf{ip}, \textbf{ethernet}, \textbf{protocol},\\  \textbf{remote}, \textbf{telephone}\end{tabular} \\
enabled &
  \begin{tabular}[c]{@{}l@{}}\textbf{enable}, enter, ui, operation,\\  guide\end{tabular} &
  \begin{tabular}[c]{@{}l@{}}\textbf{connected}, \textbf{enable}, device, \\ \textbf{configured}, setting\end{tabular} \\
change &
  one, call, time, \textbf{switch}, click &
  enter, enable, \textbf{new}, set, access \\
click &
  \begin{tabular}[c]{@{}l@{}}change, call, check, view, \\ time\end{tabular} &
  \textbf{press}, \textbf{key}, \textbf{button}, enable, ip \\
button &
  \begin{tabular}[c]{@{}l@{}}\textbf{phone}, local, may, figure,\\  \textbf{switch}\end{tabular} &
  \begin{tabular}[c]{@{}l@{}}\textbf{click}, \textbf{key}, \textbf{remote}, displays,\\  \textbf{router}\end{tabular} \\
figure &
  \begin{tabular}[c]{@{}l@{}}button, \textbf{table}, may, local, \\ unit\end{tabular} &
  \textbf{data}, \textbf{example}, \textbf{see}, \textbf{line}, guide \\\hline
\end{tabular}%
}
\caption{5 nearest neighbors for domain specific words, where the words are represented as the output given by the last hidden layer of either RoBERTa from \cite{roberta} or RoBERTa pre-trained on the corpus of E-Manuals, further compressed into a 3-D vector using PCA \cite{pca}. For each word, most related neighbours are highlighted in \textbf{bold}}
\vspace{-6mm}
\label{tab:word_embeds}
\end{table}


\subsection{A Multi-Task Learning Approach for SR and AR} \label{novel_mtl}

In our MTL framework, SR and AR models are sequential classification networks that consist of a RoBERTa encoder followed by a task-specific classification layer.  The objective of the SR model is to retrieve the section which is most relevant to the question. 
The objective of the AR model is to retrieve the answer to the question from that section. For this, we use two settings - sentence-wise and token-wise classification. 

Both SR and AR branches share the feature extraction layers of the "RoBERTa" architecture. It is well known that such a `hard parameter sharing' approach \cite{hard_parameter_sharing}  greatly reduces the problem of overfitting. Each branch has a task-specific (here task refers to one of SR or AR) binary classification layer at the end, where the output is $2$ dimensional for the SR as well as the sentence-wise AR, whereas, the output has a dimension of $(n_{t} \times 2)$ in case of the token-wise AR, where $n_t$ represents the number of tokens in the input section. 

Our architecture used has similarity with \newcite{retrieveread}; however, ours is an improved shared transformer architecture with self-attention and skip connections~\cite{vaswani2017attention}, as compared to their shared Long Short-Term Memory (LSTM) \cite{lstm} layers. 
Also, we predict non-contiguous sentences and non-contiguous spans, which makes the task difficult due to the need for detecting long-range dependencies, and thus improves the answer retrieval as compared to \newcite{retrieveread}. The underlying domain-specific pre-training of RoBERTa provides the architecture the necessary boost to capture such difficult constraints. 

\textbf{Training: }Given a question, we perform the following feed-forward approach for each section retrieved by the unsupervised IR method. 
During sentence-wise classification, the AR model takes the question, and a sentence from the current section as input, and the SR model takes the question and the current section as input. Whereas, during token-wise classification, the AR and SR models both take the question and the current section as input. The targets are to set to 1 or 0 as per the relevance of the sentences/tokens.
During backpropagation, the multi-task loss $L_{MT}$ is the average of the loss for SR and AR (similar to \newcite{ernie2.0}).

\section{Experiments and Results}
\label{experiments-results}
To assess the efficiency of EMQAP, we first evaluate the performance of the unsupervised retrieval algorithm followed by the MTL Framework on the datasets specifically curated  in Sections~\ref{qna-datasets} --~\ref{dataset_other_emanuals}. The experimental results of unsupervised algorithm is \tb{{\em detailed in suppl.}} We found that the proposed algorithm {\bf TF-IDF + T5}  performs the best. 


\subsection{ Experimental Setup} We set the unsupervised IR method to {\bf TF-IDF + T5}.  Also, we take top $K = 10$ sections retrieved given a question as input to the supervised method, since one achieves almost
$94\%$ HIT when the top-$10$ retrieved sections are considered.   
The MTL network fine-tunes the pretrained model using the S10 dataset. 
The fine-tuning is done with a batch size of $32$, and early stopping is applied using the validation loss. 
The Samsung S10 dataset, which consists of $904$ question-answer pairs with $303$ paraphrased question pairs  is divided into three sets - $634$ samples in the training set, $180$ samples in the validation set, and $90$ samples in the test set. The division ensures the paraphrased questions all fall in the same set.  [The test datasets are a bit different in Sec. \ref{eval_emqap_cqa} and  Sec. \ref{general}.]

\subsection{ Metrics} We use the following metrics for evaluation of the {\bf MTL} framework.  
 \textbf{(a). Exact Match} - Fraction of times the  predicted answer and ground truth exactly match.
 \textbf{(b). ROUGE-L \cite{lin-2004-rouge}} - F-measure metric designed for evaluation of translation and summarization. It is evaluated based on the longest common subsequence (LCS) between the actual answer and the answer predicted by a question-answering method.
   \textbf{(c). Sentence and Word Mover Similarity \cite{clark2019sentence}} - In the case of the S+WMS metric, the GloVe word embeddings~\cite{GloVe} are weighted by the word frequencies, and the sentence embeddings (obtained by averaging the GloVe word Embeddings) are weighted by the sentence lengths, and a bag of words and sentence embeddings is created. To obtain the similarity value, a linear programming solution is used to measure the distance a predicted answer's embedding has to be moved to match the actual answer. 

\subsection{Evaluating MTL framework}
\label{subsection:MTLEv}

\noindent{\bf Baselines:}
{We compare EMQAP with other baselines such as}\\
(A)  \textit{Method based on efficient passage retrieval}
\noindent \textbf{Dense Passage Retrieval (DPR)}~\cite{dpr}: A dual BERT~\cite{bert} encoder framework is used for retrieving relevant sections, and after retrieving the relevant sections, it assigns a passage selection score to each passage. Finally, a span selection method selects the span from  the section with the highest score as the answer. We fine-tune the dual-encoder framework and the span selector on our dataset.


\noindent (B)  \textit{Methods with efficient answer retrieval}

\noindent \textbf{Technical Answer Prediction (TAP)}~\cite{techqa}: \textbf{TAP} uses a cascaded architecture, where a \textbf{document ranker} ranks the top documents (here, sections) according to an assigned score, and the section with the highest score is passed to a \textbf{span selector}, which predicts the answer span. This baseline is of significance, as it has been used for the TechQA Dataset, which is the closest  to our dataset in terms of the domain.. Both the \textbf{document ranker} and the \textbf{span selector} are based on the \textbf{BERT-BASE-UNCASED} architecture, and we fine-tune both of these on S10 QA training dataset. 

\noindent \textbf{MultiSpan} \cite{simple_multi_span}: This method solves Question Answering using a sequence tagger based on the RoBERTa~\cite{roberta} architecture (we use RoBERTa-BASE architecture, as opposed to RoBERTa-LARGE as mentioned in the paper). It predicts for each token  whether it is part of the answer. For a question, the most relevant section is extracted using an IR method, and the sequence tagger is then fine-tuned using our QA Dataset. This method is of significance, as it predicts multiple spans as the answer, which matches the nature of our QA dataset.


\noindent{\bf Results:} 
Table~\ref{tab:paper_baselines} enlists the exact match, ROUGE-L precision, recall, F1 and S+WMS scores  of these baselines, along with those of sentence-wise and token-wise classification version of EMQAP.  \textbf{MultiSpan} has the highest ROUGE-L precision, and \textbf{EMQAP-S} is a close second. \textbf{TAP} is the best baseline when ROUGE-L F1 Scores and S+WMS scores are compared. However, \textbf{EMQAP-S} and \textbf{EMQAP-T} perform significantly better than \textbf{TAP}, both having p-values of approx. $0.029$. EMQAP beats all baselines, when it comes to exact match (almost no algorithm could retrieve even a single exact ground truth), S+WMS, ROUGE-L recall and F1-Scores for the following reasons - (1) The \textbf{DPR} method, although having an efficient passage retrieval, cannot select multiple spans.  
(2) Although \textbf{TAP} performs well on TechQA Dataset, it performs inferior to our method, as it cannot handle multiple spans. However, it performs better than other baselines overall, as it can give a long span as an answer, by splitting a document/section into two inputs, and later concatenating the $<START>$ token representations (3) Although \textbf{MultiSpan} can extract multiple spans as answers from a section, answer spans present in our dataset have many tokens, which  could not be handled by a Sequence Tagging Method, hence giving high ROUGE-L precision, but poor metrics otherwise. \textbf{DPR} and \textbf{MultiSpan} tend to predict very short answers, which can explain their low recall. We \tb{present} examples of different question types and their predictions by the baselines along with  ground truths in \tb{the {\em suppl.}} 

\begin{table}[H]
\centering
\resizebox{0.5\textwidth}{!}{%
\begin{tabular}{l|lllll}
\textbf{MODEL} & \textbf{EM} & \textbf{P} & \textbf{R} & \textbf{F1} & \textbf{S+WMS}\\ \hline
DPR     & 0       & 0.646      & 0.174      & 0.256 & 0.021       \\
TAP & 0.133    & 0.448           &    0.466        &  0.426           & 0.284\\
MultiSpan & 0      &  \textbf{0.938}          &  0.14          & 0.226  & 0.014          \\
\textbf{EMQAP-T} & 0.156                                                 & 0.577              & 0.682  & 0.588  & 0.34 \\  
\textbf{EMQAP-S} & \textbf{0.311} & 0.801      & \textbf{0.541}      & \textbf{0.604}    & \textbf{0.354}  
\end{tabular}%
}
\caption{Comparison of state-of-the-art models with EMQAP. (EMQAP-S and EMQAP-T are the Sentence-Wise and Token-Wise Classification variants, respectively)}
\label{tab:paper_baselines}
\end{table}

\subsection{Evaluating Pretraining techniques}
\label{mtl-experiments}

The pretrained model can be trained with different  learning rates and decay. 
Here we consider 
    \textbf{(a). FT RB}: Fine-Tuning RoBERTa~\cite{roberta}
    \textbf{(b). SLR (Same Learning Rate)}: pre-train RoBERTa on E-Manuals with Learning Rate of $5\times10^{-5}$ across all layers
    \textbf{(c). LRD (Learning Rate Decay)}: pre-train RoBERTa on E-Manuals with Learning Rate decaying linearly across layers by a factor of $2.6$, the maximum learning rate being $5\times10^{-4}$.
    \textbf{(d). EWC}: pre-train RoBERTa on E-Manuals with Elastic Weight Consolidation (EWC)
    \textbf{(e). EWC+LRD}: Combination of \textbf{EWC} and \textbf{LRD}. The strategies $c,d,$ and $e$ have been discussed in detail in Section~\ref{pretrain-strategies}. 
    Note as mentioned in Section~\ref{pretrain-strategies} EMQAP uses  \textbf{EWC+LRD}.

The efficacy of each of the  pre-trained model can be evaluated from the performance in QA system. To solely concentrate on the pre-training performance, we consider a sequential model SQP (instead of MTL) where an SR system is followed by an AR system, and each system is trained separately. Both the SR and the AR architectures are the same as that of the SR and AR branches of the MTL framework described in Section \ref{novel_mtl}.


\begin{table*}[!ht]
\centering
\small
\begin{tabular}{l|ccccc|ccccc}
\hline
\multicolumn{1}{l}{} & \multicolumn{5}{c}{\textbf{Sentence-Wise Classification}} & \multicolumn{5}{c}{\textbf{Token-Wise Classification}}                                                                                                                                                                                                                                                    \\ \hline
\multicolumn{1}{c|}{\textbf{MODEL}}                                                                                                      & \textbf{\begin{tabular}[c]{@{}c@{}}EM\end{tabular}} & \textbf{P} & \textbf{R} & \textbf{F1}   & \textbf{S+WMS} & \textbf{\begin{tabular}[c]{@{}c@{}}EM\end{tabular}} & \textbf{P} & \textbf{R} & \textbf{F1} & \textbf{S+WMS} \\ \hline
\multicolumn{1}{l|}{\begin{tabular}[c]{@{}l@{}}SQP(FT RB)\end{tabular}}                                             & 0.178                                                          & 0.696              & 0.457           & 0.506                    & 0.273                                                    &0.133*& \textbf{0.59}      & 0.602           & 0.566                    & 0.335   \\ \hline
 \multicolumn{1}{l|}{\begin{tabular}[c]{@{}l@{}}SQP(SLR)\end{tabular}}             & 0.156                                                          & 0.733              & 0.473           & 0.522                    & 0.246                  & 0.033                                                          &0.587*& 0.668           & 0.579          & 0.302       \\ \hline
 \multicolumn{1}{l|}{\begin{tabular}[c]{@{}l@{}}SQP(LRD)\end{tabular}} & 0.256                                                          & 0.783              & 0.507           & 0.57           & 0.321       & 0.089                                                          & 0.559              & 0.603           & 0.539          & 0.295     \\ \hline
 \multicolumn{1}{l|}{\begin{tabular}[c]{@{}l@{}}SQP(EWC)\end{tabular}} & 0.233                                                          & 0.763              & 0.511           & 0.552         & 0.285        & 0.1                                                            & 0.554              & 0.634           & 0.575          & 0.314    \\ \hline
 \multicolumn{1}{l|}{\begin{tabular}[c]{@{}l@{}}SQP(EWC+LRD)\end{tabular}}               &0.278*&0.791*&0.523*&0.592*&0.33*                 &0.133*& 0.574              &0.673*&0.583*&0.337* \\ \hline
   \multicolumn{1}{l|}{\begin{tabular}[c]{@{}l@{}}EMQAP\end{tabular}}               & \textbf{0.311}                                                 & \textbf{0.801}              & \textbf{0.541}  & \textbf{0.604}  & \textbf{0.354}                  & \textbf{0.156}                                                 & 0.577              & \textbf{0.682}  & \textbf{0.588}  & \textbf{0.34}\\ \hline
\end{tabular}
\caption{QA Evaluation on S10. "TF-IDF+T5" is applied by all the listed methods to select the top-10 relevant sections per question. EM stands for fraction of Exact Match. P(Precision), R(Recall) and F1 scores correspond to ROUGE-L~\cite{lin-2004-rouge}. Best result for each metric is in \textbf{bold}, while the second best is marked with $^*$} 
\label{results-1}
\end{table*}

\noindent {\bf Results:} The results are shown in Table~\ref{results-1}. Among the sentence-wise and the token-wise classification variants, the \textbf{SQP(EWC+LRD)} gives the best results considering exact match, ROUGE-L F1 and S+WMS scores, while the SQP(SLR) and the SQP(FT RB) variants perform the poorest among the lot, which is consistent with the results in~\newcite{empiricalmodelswitch}. It only produces short answers, hence have a high precision but is poor on all other counts.
Also important to note that each EWC and LRD contribute to the improvement in performance as performance of SQP with either EWC or LRD is inferior than when combined. Thus the result provides justification of using \textbf{EWC+LRD} for EMQAP.

\noindent{\bf Results: MTL over sequential learning:}
EMQAP using the \textbf{EWC+LRD} pre-training technique performs better than the best variant in all these three metric values compared to the respective sentence/token-wise classification regime. Overall, EMQAP performs better than best variant significantly with a p-value of $0.047$. 
Also, the sentence-wise model gives a higher precision, while a token-wise model gives a higher recall. This could be attributed to the sentence-wise model, in general, giving a subset of the ground truth, while the token-wise model predicting more tokens than were in the ground truth. Another metric in which sentence-wise models perform better than Token-wise classification models is Exact Match, as the token-wise models tend to miss out on some tokens in each sentence of the predicted answer. 
We \tb{present} examples of different question types and their predictions by the variants along with  ground truths in \tb{the {\em suppl.}} 

\begin{table}[H]
\small
\centering
\begin{tabular}{|c|c|c|c|c|c}
\hline
\textbf{GT} & \textbf{EM} & \textbf{P} & \textbf{R} & \textbf{F1} & \multicolumn{1}{c|}{\textbf{S+WMS}} \\ \hline
AGT & 0.304 & 0.778 & 0.522 & 0.582 & \multicolumn{1}{c|}{0.332} \\ \hline
CGT & 0.049 & 0.362 & 0.297 & 0.306 & \multicolumn{1}{c|}{0.278} \\ \hline
\end{tabular}
\caption{QA Evaluation on questions from CQA against corresponding answers from E-Manual of Samsung Smart TV as well as CQA. AGT is short for ANN-GT and CGT is short for CQA-GT ("TF-IDF+T5" is applied before all of the listed methods to select the top-10 relevant sections per question)}
\label{results-2}
\end{table}

\subsection{Evaluating Smart TV annotated on CQA Forums}
\label{eval_emqap_cqa}


We use the CQ-AQ Paraphrase dataset described in Section~\ref{reallife}. 
The $1028$ pairs of answerable questions 
and corresponding annotated answers from the manual (ANN-GT) and answers from CQA Forums (CQA-GT) are used to evaluate EMQAP. 

\noindent{\textbf{Results :}} The results obtained are tabulated in Table~\ref{results-2}. It is found that the results obtained on ANN-GT of Smart TV is inferior to that obtained on tested on S10 in Table \ref{results-1}. This happens because EMQAP is specifically fine-tuned on S10. However, we find that the performance deteriorates only a bit, pointing to the versatility of the fine-tuning. 

It is found that the Exact Match and ROUGE-L F1-Scores are not as good for the ground truths of CQA-GT as compared to ANN-GT, which could be due to different kinds of n-grams present in CQA-GT and ANN-GT, as CQA-GT has a lot of personal opinions from users in addition to the actual solution to the problem being posed in the question, while, ANN-GT, being annotated from the E-Manual, is more impersonal and informative.  However, the Mover Similarity Metrics for ANN-GT and CQA-GT are comparable which suggests  that ANN-GT and CQA-GT are semantically similar. Hence, the  Forum data can also act as a good ground truth, which we use in the next experiment.

\subsection{Evaluation  on several devices} 
\label{general}
  EMQAP is evaluated on the set of $10$ annotated questions for each device,  the details of which are provided in Section~\ref{dataset_other_emanuals}. 
  The averaged S+WMS Scores for the $4$ categories (here, sentence-wise classification is used) are tabulated in Table~\ref{other_devices}.   The mobile phones and tablets give similar results, as they have similar functionalities as S10, whereas smartwatches do not fair as well, as their functionalities differ from that of S10. 
  SQP(EWC+LRD) performance is inferior reiterating the importance of MTL. 

\begin{table}[H]
\scriptsize
\centering
\begin{tabular}{|c|c|c|c|c|}
\hline
\begin{tabular}[c]{@{}c@{}}\textbf{Sentence}\\\textbf{Wise}\\\textbf{Classification}\end{tabular} &
  \begin{tabular}[c]{@{}c@{}}Samsung\\ Galaxy\\ Mobile\\ Phones\end{tabular} &
  \begin{tabular}[c]{@{}c@{}}Other\\ Samsung\\ Mobile\\ Phones\end{tabular} &
  \begin{tabular}[c]{@{}c@{}}Samsung\\ Tablets\end{tabular} &
  \begin{tabular}[c]{@{}c@{}}Samsung\\ Smart\\Watches\end{tabular} \\ \hline
MTL (EMQAP) & \textbf{0.282} & \textbf{0.275} & \textbf{0.265} & \textbf{0.213} \\ \hline
SQP(EWC+LRD)       &0.264&0.261&0.255&0.206\\ \hline
\end{tabular}
\caption{Average S+WMS scores on CQA Forum for $4$ categories across $40$ devices for EMQAP and variants, fine-tuned on S10 dataset. Best result for each category is in \textbf{bold}, while the second best is marked with $^*$}
\label{other_devices}
\end{table}

\section{Conclusion}
In this paper, we worked on a far less studied problem of question answering from E-Manuals. 
In order to work the subject, a pre-condition was to create benchmark datasets which we painstakingly developed. We created a large corpus from E-manuals which was used in pre-training a RoBERTa architecture. This in turn  helped in developing 
a domain-specific natural language understanding; the fruits of which can be observed in the huge improvement in performance  over competing baselines.  
We believe that the E-manuals specific QA dataset is extensive and well-rounded and will help the community in various ways.  

\section*{Acknowledgements}
We would like to thank the annotators who made the curation of the datasets possible. Also, special thanks to Manav Kapadnis, an Undergraduate Student of Indian Institute of Technology Kharagpur, for his contribution towards the implementation of certain baselines. This work is supported in part by the Federal Ministry of Education and Research (BMBF), Germany under the project LeibnizKILabor (grant no. 01DD20003). This work is also supported in part by Confederation of Indian Industry (CII) and the Science \& Engineering Research Board Department of Science \& Technology Government of India (SERB) through the Prime Minister's Research Fellowship scheme. Finally, we acknowledge the funding received from Samsung Research Institute, Delhi for the work.
\bibliographystyle{acl_natbib}
\bibliography{custom}

\renewcommand\thesection{\arabic{section}}
\renewcommand\thesubsection{\thesection.\arabic{subsection}}

\appendix

\onecolumn

\section*{Supplementary Material}

\section{Introduction}
The supplementary is organized in the same sectional format as the main paper. The additional material of  a section is put in the corresponding section of the supplementary so that it becomes easier for the reader to find the relevant information. 

Some sections and subsections may not have supplementary so only their name is mentioned. 
\section{Corpus and Datasets}

\subsection{ Creating the E-Manuals corpus used for pre-training
}

\noindent \textbf{Pre-processing of Pre-training Corpus:} 
Each PDF is read in a hierarchical manner (PDF $\rightarrow$ block $\rightarrow$ span) to keep the order of the text intact, and the images are ignored (if any). The `PyMuPDF'\footnote{\url{https://pypi.org/project/PyMuPDF/}} python package is used for reading the PDFs. We remove the table of contents and all the non-Unicode and non-ASCII characters from the E-manuals. We concatenate the cleaned text of all the E-Manuals, thus collecting a total of $11,653,755$ paragraphs, each having an average of $4.4$ sentences.

\noindent \textbf{Sample paragraph from Pre-training Corpus}
Two sample paragraphs from the corpus are as follows (these samples show that the text in the corpus is mostly instructional) -

\texttt{
“1. While the printer is idle, press the Help pages menu item. 2. Note the IP address on the print and save the print for later reference. Leave the printer plugged into its power outlet; this preserves a ground path for static discharges. Touch the printer's bare metal frame often to discharge static electricity from your body. Handle the circuit board(s) by their edges only. Do not lay the board(s) on a metal surface. Make the least possible movements to avoid generating static electricity. Avoid wearing wool, nylon or polyester clothing; they generate static electricity.”\\}

\texttt{
“Batteries Warning Batteries should never be exposed to flame, heated, short-circuited or disassembled. Do not attempt to recharge alkaline, lithium or any other non-rechargeable batteries. Never use any battery with a torn or cracked outer cover. Keep batteries out of the reach of children. If you notice anything unusual when using this product such as abnormal noise, heat, smoke, or a burning odor: 1 remove the batteries immediately while being careful not to burn yourself, and; 2 call your dealer or local Olympus representative for service. AC Adapter”
}

\noindent \textbf{Word Cloud characterizing pre-train corpus}

\begin{figure}[H]
    \centering
    \includegraphics[width=0.5\linewidth]{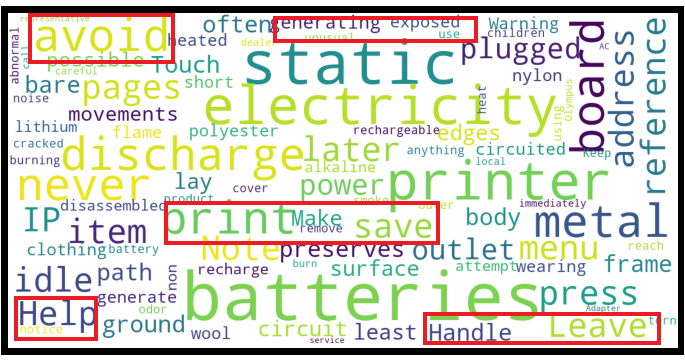}
    \caption{Word Cloud for the most frequently occurring terms in sampled paragraphs. The red boxes enclose verbs containing the instructional and assertive nature of the sentences. eg: "avoid", "help", "handle", "leave", "print".}
    \label{fig:wordcloud}
\end{figure}

Fig. \ref{fig:wordcloud} shows a word cloud for the top $200$ most frequently occurring words in the above two paragraph samples. Red boxes enclose verbs that bring out the instructional and assertive nature of the sentences.

\subsection*{Question Answering Dataset from E-Manual}

Samsung brand was chosen for the following reasons - (1) Samsung manufactures a variety of models of smartphones, Televisions and other electronic goods. (2) E-Manuals of Samsung are easily available\footnote{\url{https://www.samsung.com/us/support/downloads/}} in HTML as well as PDF formats, and are very well organized. (3) A large number of user forum questions are available in Amazon, which makes study of consumer question forums possible. However, other popular brands could also be chosen, we believe that would not make much of a difference.  (4) According to Gartner, Samsung is ranked $\#1$ in terms of Digital IQ~\footnote{\url{https://www.gartner.com/en/marketing/insights/daily-insights/top-10-consumer-electronics-brands-in-digital-3}}
\footnote{\url{https://www.gartner.com/en/marketing/insights/daily-insights/top-10-consumer-electronics-brands-in\%2Ddigital-4}}
\footnote{\url{https://www.gartner.com/en/marketing/research/digital-iq-index-consumer-electronics-us-2020}}, which may be treated as a proxy of how a brand is able to integrate in the smart technology ecosystem.

\subsection*{Analyzing quality of Annotated Question Answering Dataset} 

\begin{figure*}[h]
    \centering
    \includegraphics[width=0.9\textwidth]{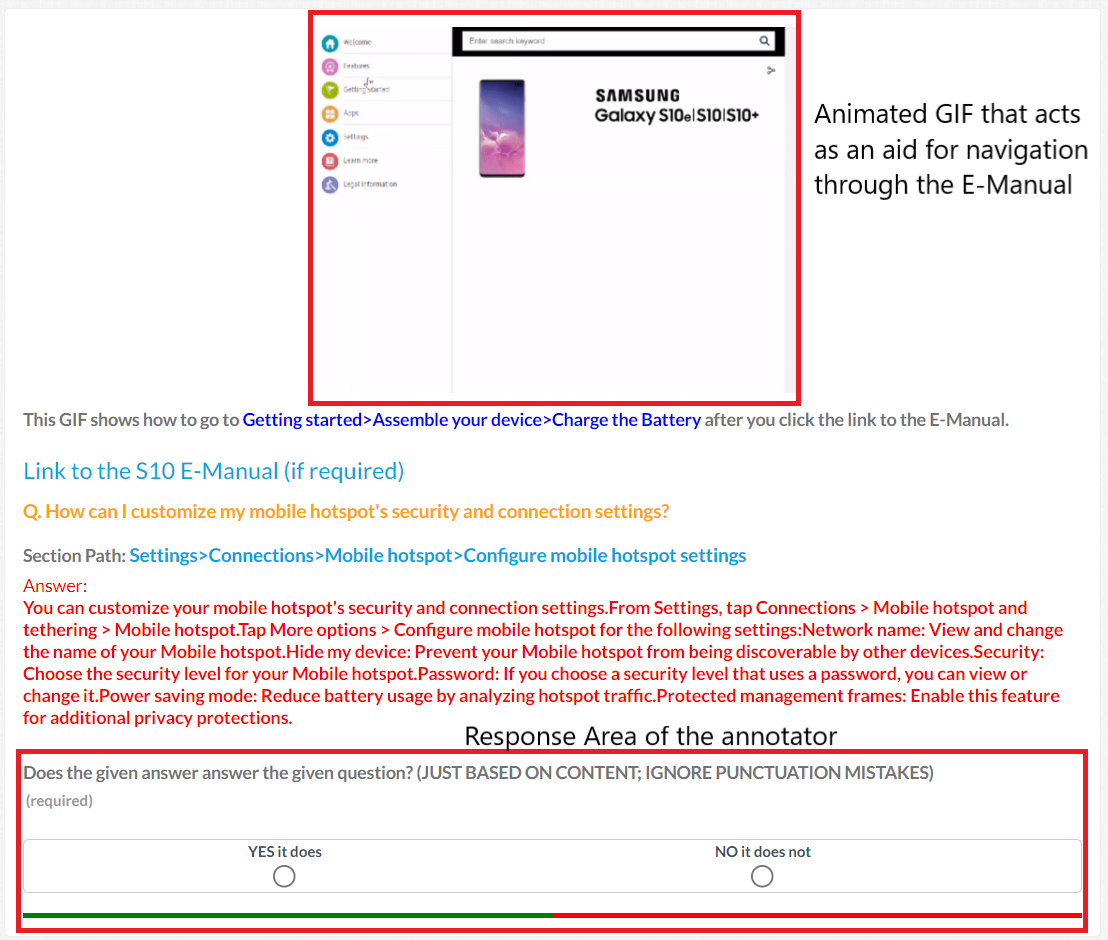}
    \caption{User Interface for the crowdworker}
    \label{fig:my_label}
\end{figure*}

In order to evaluate the quality of the expert annotations, we use the crowdsourcing platform Appen~\footnote{\url{https://client.appen.com/}} to launch two crowdsource surveys - one of S10 QA Dataset and other for Smart TV/Remote QA. $100$ QA pairs each are randomly sampled from the S10 QA and the Smart TV/Remote QA Dataset separately, and corresponding to these pairs, questions, sections containing their answers, the answers annotated by the expert annotators and the E-Manual are given to crowdworkers. Each worker needs to decide if the annotated answer satisfactorily answers the corresponding question. $3$ judgements are considered per question, and the workers that finish an annotation in less than $3$ minutes are flagged, thus avoiding spam. The crowdworkers answer using an interface illustrated in Fig. \ref{fig:my_label}.

Also, there are three levels of crowdworkers mentioned in Appen -
\textbf{Level 1} - Fastest Throughput: All qualified contributors
\textbf{Level 2} - Higher Quality: Smaller group of more experienced, higher accuracy contributors
\textbf{Level 3} - Highest Quality: Smallest group of most experienced, highest accuracy contributors
We select the `Level 3' of crowdworkers to ensure that the annotation quality of the crowdworkers performing the survey is not compromised with. Table \ref{tab:survey} shows the results of the survey, showing that for more than 95\% of the samples, majority of crowdworkers agree with the expert annotation in both the surveys. Also, the quality of the surveys in terms of clarity and ease of job is quite good based on the ratings given by some crowdworkers.

\begin{table}[H]
\centering
\begin{tabular}{lcc}
\multicolumn{1}{l}{\textbf{\begin{tabular}[c]{@{}l@{}}Measure of the agreement between \\ crowdworkers and experts\end{tabular}}}      & S10 &  Smart TV/Remote   \\ \hline
\multicolumn{1}{l|}{\begin{tabular}[c]{@{}l@{}}No. of crowdworkers (excluding flagged\\ ones)\end{tabular}}                        & 116 & 210   \\
\multicolumn{1}{l|}{\begin{tabular}[c]{@{}l@{}}No. of randomly chosen samples\\ from the S10 QA Dataset\end{tabular}}              & 100 & 100  \\
\multicolumn{1}{l|}{\begin{tabular}[c]{@{}l@{}}No. of samples where all crowdworkers\\ agree with each other and the expert\end{tabular}} & 73 & 76 \\
\multicolumn{1}{l|}{\begin{tabular}[c]{@{}l@{}}No. of samples where majority of\\ crowdworkers agree with the expert\end{tabular}} & 96   & 100 \\ \hline
\multicolumn{1}{l}{\textbf{\begin{tabular}[c]{@{}l@{}}Quality of the crowdsource survey as \\ rated by some crowdworkers\end{tabular}}}  & S10 &  Smart TV/Remote  \\ \hline
\multicolumn{1}{l|}{No. of crowdworkers who rated}                                                                                 & 13   & 8 \\
\multicolumn{1}{l|}{Average rating for clarity}                                                                                    & 3.6/5 & 4.5/5 \\
\multicolumn{1}{l|}{Average rating for ease of job}                                                                                & 3.3/5 & 4.3/5
\end{tabular}
\caption{Results of the crowdsource survey}
\label{tab:survey}
\end{table}

\subsection*{Comparison between TechQA and S10}
\begin{figure}[H]
    \centering
    \includegraphics[width = 0.4\textwidth]{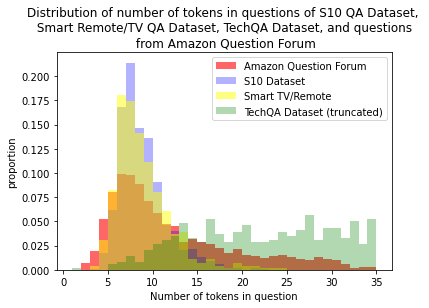}
    \caption{Comparison of normalized distributions of tokens per question of S10 QA Dataset and a set of questions extracted from Amazon Question Answering Forum}
    \label{fig:dist_comp}
\end{figure}

The size of our datasets is comparable to that of the TechQA Dataset (which belongs to the Technical Support Domain and hardly contains questions pertaining to electronics consumer products). Our datasets have \textbf{small question lengths, long answer lengths and answers that have multiple spans}, which makes it different from TechQA dataset. Also, the distribution of the number of tokens per question in our datasets is similar to that of a set of $1028$ Questions extracted from Amazon Question Answer Forum when comparing the range (approx. $5-15$) that comprises most of the density, as can be seen in Fig.~\ref{fig:dist_comp},  thus making our annotated datasets a suitable proxy for Consumer Question Answering Forums. However, a significant portion of the distribution of the question lengths in TechQA Dataset is spread over a larger range (hence truncated in Fig.~\ref{fig:dist_comp}), and is very different as compared to that of Amazon Question Answering Forum. If we consider the way that the domain-specific TechQA Dataset was curated, the questions were taken from technical forums, and answers from technical documents. However, we ask annotators to frame questions themselves from E-Manuals, by marking the answer first, and then framing the question. This would make the question set more answerable, and the questions thus obtained would be of better quality. 

\subsection{Questions from the real consumers}

\subsection{Questions spanning across several devices}

\subsection*{Sample Questions for analysis on other devices }These are the $10$ sample questions that were asked across several devices -

\begin{enumerate}
\item Does it use a sim card? 
\item How do I switch off the device?
\item Does it use a SD port?
\item Does this device offer Wi-Fi calling ?
\item How can I change the device language ?
\item How can I set the brightness level ?
\item How can I hide the notifications ?
\item How can I change the Font size ?
\item How can I use stopwatch? 
\item How do I setup tones on my device?
\end{enumerate}

\begin{figure}[H]
    \includegraphics[width=0.5\textwidth]{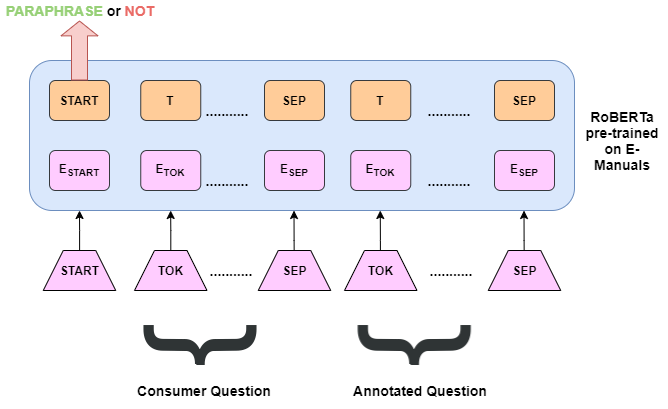}
    \caption{Question Paraphrase Detector}
    \label{fig:q_paraphrase}
\end{figure}

\noindent \textbf{Question Paraphrase Detector:} This is used for detecting Amazon User-Forum Questions that are answerable, by detecting whether it is a paraphrase of the most similar Annotated Question or not. For this, the CQ-AQ Paraphrase Dataset is split into train, validation and test sets in the ratio of $8:2:1$ for training and evaluating a \textbf{question paraphrase detector} - this is a RoBERTa Sequential Classification Model (initialized by weights of RoBERTa pre-trained on E-Manuals), as shown in Fig.~\ref{fig:q_paraphrase}. This method gives a high precision of $0.932$, and a high recall of $0.814$.

\section{Methodology}

\subsection*{Overview of Pipeline}

The EMQAP  is laid out in the form of a pseudo-code in Algorithm~\ref{algo1}.

\begin{algorithm*}[]
\SetAlgoLined
\DontPrintSemicolon
    \SetKwFunction{FMain}{Main}
    \SetKwFunction{FPreTraining}{Pre-Training}
    \SetKwFunction{FMTL}{MultiTaskLearning}
    \SetKwFunction{FIR}{InformationRetrieval}
    \SetKwFunction{FRC}{ReadingComprehension}
    \SetKwProg{Fn}{Function}{:}{}
    \Fn{\FPreTraining{$corpus, RoBERTa$}} {
    $model$ = initializeWeights($RoBERTa$, weights from \cite{roberta}) \;
    $pre\textnormal{-}trainedModel$ \textnormal{=} MaskedLanguageModeling($model$, $corpus$) \;
    \KwRet $pre\textnormal{-}trainedModel$ \;
    }
    \;
    \Fn{\FMTL{$pre\textnormal{-}trained\textnormal{-}model$, $Annotated\textnormal{-}QnA$, $E\textnormal{-}Manual$)}}{
    $copy\textnormal{-}weights$($pre\textnormal{-}trained\textnormal{-}model$.$encoder$, $supervised\textnormal{-}IR$.$encoder$) \;
    $copy\textnormal{-}weights$($pre\textnormal{-}trained\textnormal{-}model$.$encoder$, $supervised\textnormal{-}RC$.$encoder$) \;
    //batch fine-tuning \;
    \For{$QnA\textnormal{-}batch$ \textnormal{in} $Annotated\textnormal{-}QnA$}{
    $questions$, $annotated\textnormal{-}answers$ = $QnA\textnormal{-}batch$ \;
    $topK\textnormal{-}sections\textnormal{-}batch$ = [$unsupervised\textnormal{-}IR$($question$, $E\textnormal{-}Manual$) for $question$ in $questions$] \;
    $IR\textnormal{-}prediction$ = $supervised\textnormal{-}IR$($questions$, $topK\textnormal{-}sections\textnormal{-}batch$) \;
    $RC\textnormal{-}prediction$ = $supervised\textnormal{-}RC$($questions$, $topK\textnormal{-}sections\textnormal{-}batch$) \;
    $IR\textnormal{-}Loss$ = $Loss\textnormal{-}Function$($IR\textnormal{-}prediction$, $sections\textnormal{-}containing\textnormal{-}annotated\textnormal{-}answers$) \;
    $RC\textnormal{-}Loss$ = $Loss\textnormal{-}Function$($RC\textnormal{-}prediction$, $annotated\textnormal{-}answers$) \;
    $Loss$ = $average$($IR\textnormal{-}Loss$, $RC\textnormal{-}Loss$) \;
    $Back\textnormal{-}propagate$($Loss$, $supervised\textnormal{-}IR$, $supervised\textnormal{-}RC$) \;
    }
    \KwRet $supervised\textnormal{-}IR$, $supervised\textnormal{-}RC$
    }
    \;
    \Fn{\FMain{}}{
    extract $listOfEManualURLS$ from \url{www.manualsonline.com} \;
    $corpus$ = $createCorpus$($listOfEManualURLS$) \;
    $pre\textnormal{-}trainedModel$ \textnormal{=} $preTraining$($corpus$, $RoBERTa$) \;
    $supervised\textnormal{-}IR$, $supervised\textnormal{-}RC$ = $MultiTaskLearning$($pre\textnormal{-}trainedModel$, $AnnotatedQnA$, $E\textnormal{-}Manual$) \;
    \textnormal{//inference, given a question and the E-Manual from which the question is asked.} \;
    $topK\textnormal{-}sections$ \textnormal{=} $unsupervised\textnormal{-}IR$($question$, $E\textnormal{-}Manual$) \;
    $pred\textnormal{-}section$ \textnormal{=} $argmax$($supervised\textnormal{-}IR$($question$, $topK\textnormal{-}sections$)) \;
    $pred\textnormal{-}answer$ = $hard\textnormal{-}classifier$($supervised\textnormal{-}RC$($question$, $pred\textnormal{-}section$)) \;
    \KwRet $pred\textnormal{-}answer$ \;
    }
\caption{EMQAP Pipeline}
\label{algo1}
\end{algorithm*}

\subsection*{Retrieving top $k$ sections} \label{unsup_ir}

Given an E-Manual, our first step is to reduce the pipeline's search space and provide it with only a few candidate sections for a question. We use an unsupervised IR method that accepts a question and all sections of the E-Manual as input and provides similarity scores for each question-section as output. We select the $K$ highest scoring sections, which possibly contain the answer.
Experiments show that the best way of representing question-section is by TF-IDF. Thus we create TF-IDF vector representations of questions and sections and calculate the cosine similarity of each question-section pair. 

However, we make an enhancement by 
augmenting a section with probable questions that can be answered by that section  ~\cite{Nogueira2019DocumentEB}. These questions are generated by a pre-trained T5 (Text-to-Text Transfer Transformer) \cite{raffel2019exploring} model, which takes the section as input and outputs a list of questions that are answerable by that section.  This augmentation results in the re-weighting of the terms, especially the terms which act as anchor when  questions are framed  receive more weights.   We find this leads to improved retrieval of top $k$ sections. We name this improvisation as TF-IDF + T5.
\subsection{Pre-training on the E-Manuals Corpus}

{\bf State-of-the-art pre-training}
of
transformer models include masked language model pre-training \cite{bert,roberta}, next sentence prediction \cite{bert}, elastic weight consolidation (EWC) \cite{ewc}, a decaying learning rate as a function of layer depth \cite{empiricalmodelswitch}, using heuristic data selection methods for an experience replay buffer \cite{exprep}, etc. Also, domain-adaptive fine-tuning methods have been used for transformer language models pre-trained on generic data such as ELMo~\cite{elmo} and BERT~\cite{bert} in order to improve performance in downstream tasks such as sequence labelling~\cite{unsup_adapt_seq_lab}, duplicate question detection~\cite{duplicate_q_det} etc. \newcite{unsup_dom_ad} suggests unsupervised domain adaptation methods, that do not even require domain-specific annotated data.\\

\begin{figure}[h]
    \centering
    \includegraphics[width=8cm, height=3.2cm]{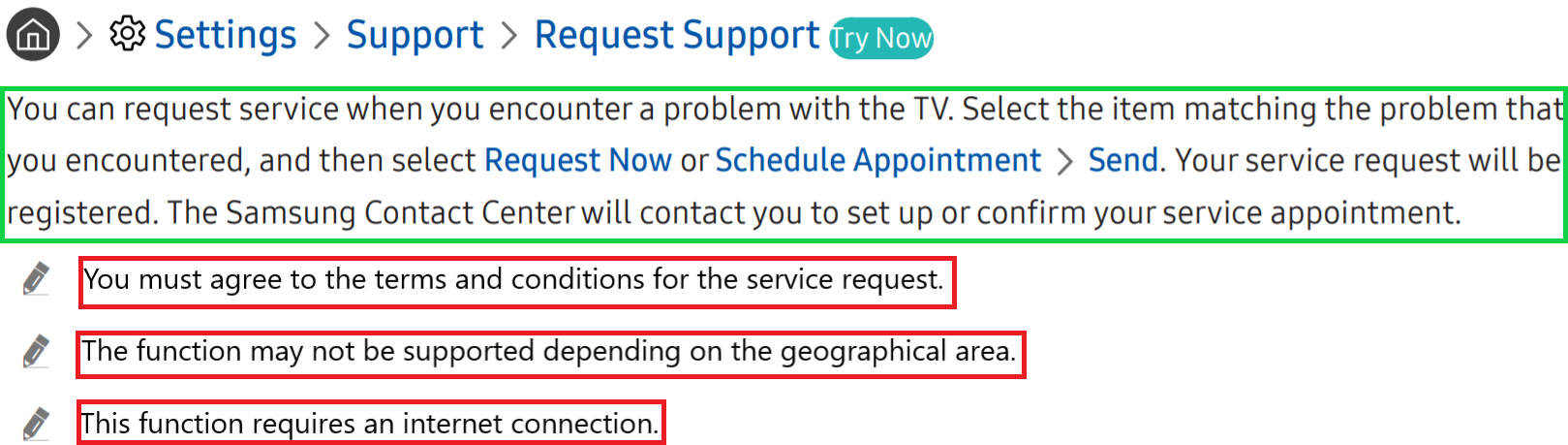}
    \caption{A sample from an E-Manual. Although the sentences enclosed by red boxes are adjacent, they are independent of each other. Instead, each such sentence is dependent on the sentences in the green box.}
    \label{fig:smart-remote-1}
\end{figure}

\noindent {\bf Justification behind using masked language modeling}
We did not use the Next Sentence Prediction (NSP) pre-training task \cite{bert}, as it has been shown in \newcite{roberta,xlnet,spanbert} that NSP worsens performance in downstream QNLI \cite{glue} tasks and in question answering on the SQuAD Dataset \cite{squad}. Also, intuitively, sentences in E-Manuals sometimes do not have dependencies with an adjacent sentence. Instead, there might be many sentences that are dependent on a particular statement that is not necessarily adjacent, as shown in Fig.~\ref{fig:smart-remote-1}. \\ 

\noindent {\bf Justification behind having a single epoch iteration.}
We pre-train RoBERTa on E-Manuals only for $1$ epoch. This is as per the justifications put forward by ~\newcite{one_epoch}.  (1) Single epoch ensures better diversity in the samples processed as compared to multi-epoch training thus preventing overfitting (2) Sampling from the training data matches the underlying data distribution in single epoch (3). RoBERTa has about $125M$ parameters. 
In our case, the number of batches is close to $80,000$, and the number of tokens (T) in the pre-training E-Manuals corpus is close to $1B$, making the ratio $T/P \approx 8$, which satisfies the optimal conditions for pre-training for one epoch as per ~\newcite{one_epoch}.

\subsection{A Multi-Task Learning Approach for SR and AR}

\section{Experiments and Results}

\noindent{\bf Evaluation of unsupervised IR methods}
\label{metrics}

We evaluate the performance of our algorithm  {\bf TF-IDF + T5} ({\em detailed in suppl.}) with different baselines. 

\noindent{\bf Baselines:} We evaluate several baselines, such as - 
    (a). Jaccard Similarity (\textbf{Jaccard Sim}) and Word Count Vector Similarity (\textbf{Count Vec Sim}) between the tokens of a question and the sections.
    (b). {\bf Cosine similarity} between averaged pre-trained neural word embedding vectors such as \textbf{word2vec}~\cite{word2vec}, \textbf{GloVe}~\cite{GloVe}, and \textbf{FastText}~\cite{fastText} of the tokens of a question and the sections.
    (c). {\bf Cosine similarity} between the sparse vectors generated using \textbf{TF-IDF} on tokens of a question and the sections.    
    (d). {\bf Cosine similarity} between pre-trained neural sentence vectors like \textbf{InferSent}~\cite{InferSent}  of a question and the sections.


\begin{table}[!htb]
\centering
\begin{tabular}{|c|c|c|c|}
\hline
                       & \textbf{Hits@1} & \textbf{Hits@5} & \textbf{Hits@10} \\ \hline
\textbf{InferSent} & 0.033           & 0.1           & 0.156            \\ \hline
\textbf{Jaccarrd Sim} & 0.222           & 0.422           & 0.467            \\ \hline
\textbf{Count Vec Sim}        & 0.333           & 0.6           & 0.633            \\ \hline
\textbf{GloVe Sim}   & 0.256  & 0.567  & 0.711   \\ \hline
\textbf{fasttext\_sim} & 0.356           & 0.711           & 0.756            \\ \hline
\textbf{word2vec\_sim} & 0.333           & 0.711           & 0.767            \\ \hline
\textbf{TF-IDF}        & 0.511           & 0.889           & 0.911            \\ \hline
\textbf{TF-IDF + T5}   & \textbf{0.533}  & \textbf{0.911}  & \textbf{0.934}   \\ \hline
\end{tabular}
\caption{Unsupervised Information Retrieval Methods evaluated on S10 QA.}
\label{table_unsup_IR}
\end{table}

\noindent{\bf Results:} We evaluate Hits@$K$ that is, the fraction of the number of times the section relevant to a question appears in the top $K$ sections for the baselines and
(\textbf{TF-IDF+T5}) and report the results  in the Table \ref{table_unsup_IR} for the test set of $90$ questions of the S10 QA dataset.  As can be seen  \textbf{TF-IDF+T5},   gives the best Hits@$K$ for $K =  {1, 5, 10}$ values = {$0.533$, $0.911$, $0.934$}. 
 \subsection{Evaluating MTL Framework}

Table~\ref{table:examples_1} shows three examples of questions and the corresponding predictions of EMQAP (Sentence-Wise Classification) and baselines.

\begin{table*}[!ht]
\tiny
\centering
\begin{tabular}{|c|l|l|l|}
\hline
\textbf{Question} &
  \begin{tabular}[c]{@{}l@{}}How can I turn on and turn off fast wireless charging?\end{tabular} &
  \begin{tabular}[c]{@{}l@{}}Where can I find an option to setup separate app sound?\end{tabular} &
  What is Samsung DeX for PC? \\ \hline
\textbf{\begin{tabular}[c]{@{}c@{}}Ground\\ Truth\\ Answer\end{tabular}} &
  \begin{tabular}[c]{@{}l@{}}From Settings, tap Device care > Battery \\ for options. Fast wireless charging -\\ Enable or disable fast wireless \\ charging when using a supported charger.\end{tabular} &
  \begin{tabular}[c]{@{}l@{}}You can play media sound on a speaker or headphones separate from \\ the rest of the sounds on your device. Connect to a Bluetooth device \\ to make this option available in the Audio device menu. From Settings \\ tap Sounds and vibration > Separate app sound. Tap Turn on now to \\ enable Separate app sound and then set the following options - App > \\ Choose an app to play its sound on a separate audio device. Audio \\ device - Choose the audio device that you want the app's sound \\ to be played on\end{tabular} &
  \begin{tabular}[c]{@{}l@{}}Connect your device to a PC for an enhanced \\ multitasking experience. Use your device \\ and PC apps side by side. \\ Share the keyboard mouse and screen \\ between the two devices. Make phone \\ calls or send texts while using DeX\\. samsung.com/us/explore/dex\end{tabular} \\ \hline
\textbf{EMQAP} &
  \begin{tabular}[c]{@{}l@{}}From Settings tap Device care > Battery \\ for options. Battery PowerShare - Enable wireless \\ charging of supported devices with your device's \\ battery. Fast cable charging - Enable or disable \\ fast cable charging when connected to a supported \\ charger \end{tabular} &
  \begin{tabular}[c]{@{}l@{}}You can play media sound on a speaker or headphones separate from \\ the rest of the sounds on your device. Connect to a Bluetooth device \\ to make this option available in the Audio device menu. From Settings \\ tap Sounds and vibration > Separate app sound. Tap Turn on now to \\ enable Separate app sound and then set the following options - App > \\ Choose an app to play its sound on a separate audio device. Audio \\ device - Choose the audio device that you want the app's sound \\ to be played on\end{tabular} &
  \begin{tabular}[c]{@{}l@{}}Connect your device to a PC for an enhanced \\ multitasking experience. Use your device \\ and PC apps side by side. \\ Share the keyboard mouse and screen \\ between the two devices. Make phone calls \\ or send texts while using DeX. Visit for more \\ information - samsung.com/us/explore/dex\end{tabular} \\ \hline
\textbf{\begin{tabular}[c]{@{}c@{}}DPR\end{tabular}} &
  \begin{tabular}[c]{@{}l@{}}depending on device condition\\ or surrounding environment\end{tabular} &
  \begin{tabular}[c]{@{}l@{}}Settings\end{tabular} &
  \begin{tabular}[c]{@{}l@{}}Volume. Tap More options > Media\\ volume limit\end{tabular} \\ \hline    
\textbf{\begin{tabular}[c]{@{}c@{}}MultiSpan\end{tabular}} &
  \begin{tabular}[c]{@{}l@{}}Enable \\\end{tabular} &
  \begin{tabular}[c]{@{}l@{}}Audio device menu\end{tabular} &
  \begin{tabular}[c]{@{}l@{}}enhanced, multitasking\end{tabular} \\ \hline      
\textbf{\begin{tabular}[c]{@{}c@{}}TAP\end{tabular}} &
  \begin{tabular}[c]{@{}l@{}}Select a power mode to extend battery life. App power \\management : Configure battery usage for apps that are \\used infrequently. Wireless PowerShare : Enable wireless \\charging of supported devices with your devices battery. \\Fast cable charging : Enable or disable fast cable charging\\ when connected to a supported charger. Fast wireless\\ charging : Enable or disable fast wireless \\charging when using a supported charger.\end{tabular} &
  \begin{tabular}[c]{@{}l@{}}make this option available in the Audio device\\ menu. From Settings, tap Sounds and vibration > Separate\\ app sound . Tap Turn on now to enable Separate app\\ sound, and then set the following options: App : \\Choose an app to play its sound on a separate\\ audio device. Audio device : Choose\\ the audio device that you want the apps sound to be played on.\end{tabular} &
  \begin{tabular}[c]{@{}l@{}}device to a PC for an enhanced,\\ multitasking experience. Use your device\\ and PC apps side-by-side Share the\\ keyboard, mouse, and screen between \\the two devices Make phone calls or\\ send texts while using DeX Visit \\samsung.com/us/explore/dex for more\\ information.\end{tabular} \\ \hline      
\multicolumn{1}{|l|}{\textbf{Remarks}} &
  \begin{tabular}[c]{@{}l@{}}For complex procedural questions, EMQAP and TAP \\give the answer closest to the ground truth.\end{tabular} &
  \begin{tabular}[c]{@{}l@{}}For `where' type questions, (asking the location of a particular \\ feature), EMQAP again performs very well as compared \\ to the other baselines.\end{tabular} &
  \begin{tabular}[c]{@{}l@{}}Factual (`what' type) questions are answered \\ equally well by EMQAP and TAP.\end{tabular} \\ \hline
\end{tabular}
\caption{Examples of question-answer pairs from the Samsung S10 QA Dataset and predictions by EMQAP (sentence-wise classification) and baselines with remarks, explaining the predictions.}
\label{table:examples_1}
\end{table*}

\subsection{Evaluating Pretraining Techniques}

\begin{figure}[H]
    \centering
    \includegraphics[scale=0.7]{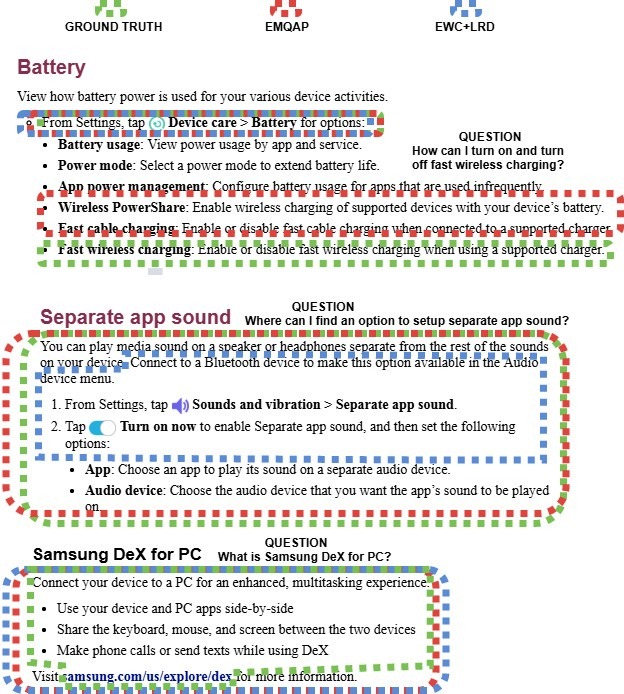}
    \caption{Ground Truth answers and the answers predicted by EMQAP and SQP(EWC+LRD)  (both using sentence-wise classification) corresponding to three questions.}
    \label{fig:QA_bbox}
\end{figure}

We present three examples of different question types and their predictions and ground truths in Table~\ref{table:examples} given by $2$ variants and EMQAP, along with some remarks. We observe that EMQAP gives better answers for questions that inquire about procedure or location compared to variants. However, factual questions are answered similarly by all the models. Also, considering  Table~\ref{fact_proc_loc}, we can see that EMQAP performs better than SQP(EWC+LRD) in all three categories, making a considerable improvement in answering location-based questions. Hence, we can say that questions regarding the device's operation and features are answered better by the EMQAP compared to all other variants. Also, the SQP(EWC+LRD) variant is better than the SQP(SLR) in answering the questions, which indicates the superiority of the training scheme. If we consider the questions containing non-contiguous ground truths, EMQAP performs better than SQP(EWC+LRD), as can be seen in Fig. \ref{fig:QA_bbox}.




\begin{table*}[!ht]
\tiny
\centering
\begin{tabular}{|c|l|l|l|}
\hline
\textbf{Question} &
  \begin{tabular}[c]{@{}l@{}}How can I turn on and turn off fast wireless charging?\end{tabular} &
  \begin{tabular}[c]{@{}l@{}}Where can I find an option to setup separate app sound?\end{tabular} &
  What is Samsung DeX for PC? \\ \hline
\textbf{\begin{tabular}[c]{@{}c@{}}Ground\\ Truth\\ Answer\end{tabular}} &
  \begin{tabular}[c]{@{}l@{}}From Settings, tap Device care > Battery \\ for options. Fast wireless charging -\\ Enable or disable fast wireless \\ charging when using a supported charger.\end{tabular} &
  \begin{tabular}[c]{@{}l@{}}You can play media sound on a speaker or headphones separate from \\ the rest of the sounds on your device. Connect to a Bluetooth device \\ to make this option available in the Audio device menu. From Settings \\ tap Sounds and vibration > Separate app sound. Tap Turn on now to \\ enable Separate app sound and then set the following options - App > \\ Choose an app to play its sound on a separate audio device. Audio \\ device - Choose the audio device that you want the app's sound \\ to be played on\end{tabular} &
  \begin{tabular}[c]{@{}l@{}}Connect your device to a PC for an enhanced \\ multitasking experience. Use your device \\ and PC apps side by side. \\ Share the keyboard mouse and screen \\ between the two devices. Make phone \\ calls or send texts while using DeX\\. samsung.com/us/explore/dex\end{tabular} \\ \hline
\textbf{EMQAP} &
  \begin{tabular}[c]{@{}l@{}}From Settings tap Device care > Battery \\ for options. Battery PowerShare - Enable wireless \\ charging of supported devices with your device's \\ battery. Fast cable charging - Enable or disable \\ fast cable charging when connected to a supported \\ charger \end{tabular} &
  \begin{tabular}[c]{@{}l@{}}You can play media sound on a speaker or headphones separate from \\ the rest of the sounds on your device. Connect to a Bluetooth device \\ to make this option available in the Audio device menu. From Settings \\ tap Sounds and vibration > Separate app sound. Tap Turn on now to \\ enable Separate app sound and then set the following options - App > \\ Choose an app to play its sound on a separate audio device. Audio \\ device - Choose the audio device that you want the app's sound \\ to be played on\end{tabular} &
  \begin{tabular}[c]{@{}l@{}}Connect your device to a PC for an enhanced \\ multitasking experience. Use your device \\ and PC apps side by side. \\ Share the keyboard mouse and screen \\ between the two devices. Make phone calls \\ or send texts while using DeX. Visit for more \\ information - samsung.com/us/explore/dex\end{tabular} \\ \hline
\textbf{\begin{tabular}[c]{@{}c@{}}SQP(EWC\\ +\\ LRD)\end{tabular}} &
  \begin{tabular}[c]{@{}l@{}}From Settings tap Device care > Battery for options.\end{tabular} &
  \begin{tabular}[c]{@{}l@{}}Connect to a Bluetooth device to make this option available in the \\ Audio device menu. From Settings tap Sounds and vibration > \\ Separate app sound. Tap Turn on now to enable Separate app sound \\ and then set the following options\end{tabular} &
  \textless{}SAME AS EMQAP\textgreater{} \\ \hline
\textbf{SQP(SLR)} &
  \begin{tabular}[c]{@{}l@{}}From Settings, tap Device care > Battery for options. \\ Battery usage - View power usage by app and service. \\ Power mode - Select a power life > App > power \\ management. Configure Power.\end{tabular} &
  From Settings tap and &
  \textless{}SAME AS EMQAP\textgreater{} \\ \hline
\multicolumn{1}{|l|}{\textbf{Remarks}} &
  \begin{tabular}[c]{@{}l@{}}For complex procedural questions, EMQAP \\give the answer closest to the ground truth.\end{tabular} &
  \begin{tabular}[c]{@{}l@{}}For `where' type questions, (asking the location of a particular \\ feature), EMQAP again performs very well as compared \\ to the other two variants.\end{tabular} &
  \begin{tabular}[c]{@{}l@{}}Factual (`what' type) questions are answered \\ equally well by EMQAP as well as the \\ variants.\end{tabular} \\ \hline
\end{tabular}
\caption{Examples of question-answer pairs from the Samsung S10 QA Dataset and predictions by EMQAP and two variants (sentence-wise classification in AR model), with remarks, explaining the predictions.}
\label{table:examples}
\end{table*}

\begin{table}[H]
\small
\centering
\begin{tabular}{|c|c|c|c|c|c}
\hline
\textbf{MODEL} & \textbf{Factual} & \textbf{Procedural} &  \multicolumn{1}{c|}{\textbf{Location}} \\ \hline
EMQAP & \textbf{0.455} & \textbf{0.582} & \multicolumn{1}{c|}{\textbf{0.664}} \\ \hline
SQP(EWC+LRD) & 0.417 & 0.576 & \multicolumn{1}{c|}{0.561} \\ \hline
\end{tabular}
\caption{Average F1-Scores for factual, procedural and location-based questions on test set of S10 QA Dataset}
\label{fact_proc_loc}
\end{table}

\begin{figure}[H]
    \centering
    \includegraphics[scale=0.7]{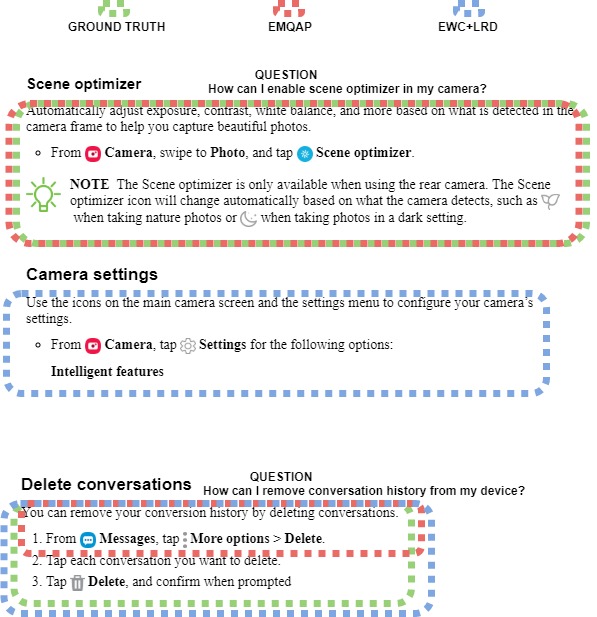}
    \caption{Ground Truth answers and the answers predicted by EMQAP and SQP(EWC+LRD) (both using sentence-wise classification) corresponding to two questions. In  the first question, SQP(EWC+LRD) selects a wrong section, while in the second question, EMQAP does not give the complete answer.}
    \label{fig:QA_bbox_2}
\end{figure}

Fig.~\ref{fig:QA_bbox} shows Ground Truth answers and the answers predicted by EMQAP and SQP(EWC+LRD) (both using sentence-wise classification) corresponding to three questions mentioned in Table \ref{table:examples}. Fig.~\ref{fig:QA_bbox_2} similarly shows two more questions, but the first question shows how SQP(EWC+LRD) selects a wrong section when the answer is long, whereas, in the second question, EMQAP does not give the complete answer, while SQP(EWC+LRD) gives the correct answer.

\section{Evaluating Smart TV annotated on CQA Forums}

\section{Evaluation on several devices}

\end{document}